    \documentclass[conference]{IEEEtran}
    
    \usepackage{tikzit}
    \usepackage{tikz}

\tikzstyle{black empty dot}=[fill=none, draw=black, shape=circle, scale=1, thick]
\tikzstyle{red dot}=[fill={rgb,255: red,255; green,127; blue,127}, draw=none, shape=circle, scale=0.3]
\tikzstyle{green dot}=[fill={rgb,255: red,127; green,201; blue,127}, draw=none, shape=circle, scale=0.3]
\tikzstyle{rectangle}=[fill=white, draw=black, shape=rectangle]
\tikzstyle{large box}=[fill=none, draw=black, shape=rectangle, minimum width=0.75cm, minimum height=1cm, rounded corners=2mm]
\tikzstyle{trapezium}=[fill={rgb,255: red,150; green,186; blue,218}, draw=none, shape=trapezium, minimum height=1cm, minimum width=1cm, rotate=270, rounded corners=0.2mm]
\tikzstyle{invisible large box}=[fill=none, draw=none, shape=rectangle, minimum width=0.75cm, minimum height=1cm]
\tikzstyle{dashed circle}=[fill=none, draw=black, shape=circle, minimum size=1.9cm, dashed]
\tikzstyle{star}=[fill=white, draw=black, shape=star, scale=0.3]

\tikzstyle{standard edge}=[draw=black, -, fill={rgb,255: red,215; green,217; blue,218}, fill opacity=0.2]
\tikzstyle{big dash}=[-, thick, dashed, dash pattern=on 4mm off 2mm, fill=none]
\tikzstyle{dashed edge}=[draw=black, -, dashed]
\tikzstyle{arrow}=[->, thick]
\tikzstyle{arrow2}=[{|->}]
\tikzstyle{arrow left}=[<-]
\tikzstyle{arrow left2}=[{<-|}]
\tikzstyle{arrow both sides}=[<->]
\tikzstyle{blue edge}=[-, draw=blue, thick]
\tikzstyle{blue arrow}=[draw=blue, ->, thick]
\tikzstyle{red edge}=[-, draw=red, thick]
\tikzstyle{dash points}=[-, dash pattern=on 0.01mm off 0.1mm, dashed]
\tikzstyle{thin edge}=[-, line width=0.5, fill={rgb,255: red,122; green,190; blue,232}, fill opacity=0.5, draw={rgb,255: red,128; green,128; blue,128}]
\tikzstyle{gray}=[-, draw={rgb,255: red,191; green,191; blue,191}]
\tikzstyle{edge}=[-, fill=white]
\tikzstyle{red pastel edge}=[-, line width=0.5, fill={rgb,255: red,255; green,129; blue,129}, fill opacity=0.5, draw={rgb,255: red,128; green,128; blue,128}]
\tikzstyle{yellow edge}=[-, line width=0.5, fill={rgb,255: red,253; green,227; blue,227}, fill opacity=0.5, draw={rgb,255: red,128; green,128; blue,128}]
\tikzstyle{new edge style 0}=[-, fill={rgb,255: red,96; green,225; blue,67}, draw={rgb,255: red,128; green,128; blue,128}, fill opacity=0.5, line width=0.5]
\tikzstyle{normal edge}=[-, thick]

    \usepackage{cite}
    \usepackage{amsmath,amssymb,amsfonts}
    \usepackage{algorithmic}
    \usepackage{graphicx}
    \usepackage{textcomp}
    \usepackage{xcolor}
    \usepackage{hyperref}
    \usepackage[utf8]{inputenc}
    \usepackage{pgfplots}
    \usepackage{subfigure}
    \DeclareUnicodeCharacter{2212}{−}
    \usepgfplotslibrary{groupplots,dateplot}
    \usetikzlibrary{patterns,shapes.arrows}
    \pgfplotsset{compat=newest}
    \def\BibTeX{{\rm B\kern-.05em{\sc i\kern-.025em b}\kern-.08em
        T\kern-.1667em\lower.7ex\hbox{E}\kern-.125emX}}
    
    \makeatletter
    \newcommand{\linebreakand}
    {
        \end{@IEEEauthorhalign}
        \hfill\mbox{}\par
        \mbox{}\hfill\begin{@IEEEauthorhalign}
    }
    \makeatother
    
    \usepackage{arydshln}
    
    \newcommand{\cstN}{N}
    
    \newcommand{\cstC}{C}
    \newcommand{\cstK}{K}
    \newcommand{\funW}{W}
    \newcommand{\varCin}{c_\mathrm{in}}
    \newcommand{\varCout}{c_\mathrm{out}}
    \newcommand{\varI}{i}
    \newcommand{\varJ}{j}
    \newcommand{\varK}{k}
    \newcommand{\varL}{l}
    \newcommand{\varV}{v}
    
    \newcommand{\varC}{c}
    \newcommand{\varH}{h}
    \newcommand{\varD}{d}
    \newcommand{\graphG}{{\mathcal G}}
    \newcommand{\setV}{{\mathcal V}}
    \newcommand{\setE}{{\mathcal E}}
    \newcommand{\vecX}{{\bf x}}
    
    \newcommand{\vecH}{{\bf h}}

    \newcommand{\vecXs}{{\bf \hat{x}}}
    
    \newcommand{\matI}[1]{{\bf I}_{#1}}
    \newcommand{\matW}{{\bf W}}
    \newcommand{\matD}{{\bf D}}
    \newcommand{\matX}{{\bf X}}
    
    \newcommand{\matXs}{{\bf \hat{x}}}
    \newcommand{\matZ}{{\bf Z}}
    \newcommand{\matL}{{\bf L}}
    \newcommand{\matU}{{\bf U}}
    \newcommand{\matLambda}{{\boldsymbol \Lambda}}
    \newcommand{\tenW}{{\boldsymbol \Theta}}
    
    \newcommand{\R}{{\mathbb R}}
    
    \newcommand{\Order}{{\mathcal O}}
    \newcommand{\entry}[2]{#1\left[ #2 \right]}
    \newcommand{\set}[1]{\left\{ #1 \right\}}
    \newcommand{\tuple}[1]{\left\langle #1 \right\rangle}
    \newcommand{\card}[1]{\left| #1 \right|}
    \DeclareMathOperator{\gft}{GFT}
    \DeclareMathOperator{\igft}{GFT^{-1}}
    \DeclareMathOperator{\gspConv}{GSPConv}

\begin{document}
    
    \title{Pruning Graph Convolutional Networks to select meaningful graph frequencies for fMRI decoding}

    \author
    {
        \IEEEauthorblockN{Yassine El Ouahidi, Hugo Tessier, Giulia Lioi, Nicolas Farrugia, Bastien Pasdeloup and Vincent Gripon}
        \IEEEauthorblockA{\textit{IMT Atlantique} \\
                          \textit{Lab-STICC, UMR CNRS 6285} \\
                          F-29238 Brest, France \\
                          \emph{name}.\emph{surname}@imt-atlantique.fr}
    }

    \maketitle
    
    
    \begin{abstract}
        Graph Signal Processing is a promising framework to manipulate brain signals as it allows to encompass the spatial dependencies between the activity in regions of interest in the brain.
        In this work, we are interested in better understanding what are the graph frequencies that are the most useful to decode fMRI signals.
        To this end, we introduce a deep learning architecture and adapt a pruning methodology to automatically identify such frequencies.
        We experiment with various datasets, architectures and graphs, and show that low graph frequencies are consistently identified as the most important for fMRI decoding, with a stronger contribution for the functional graph over the structural one.
        We believe that this work provides novel insights on how graph-based methods can be deployed to increase fMRI decoding accuracy and interpretability.
    \end{abstract}

    

    
\section{Introduction}
\label{sec:intro}

 The development of Functional Magnetic Resonance Imaging (fMRI) has allowed to observe the brain \textit{in vivo} and to address one of the central questions of cognitive neuroscience: understanding the relation between brain activation and cognitive functions or experimental conditions. Classical approaches have relied on \textit{forward inference} (\emph{e.g.}, identifying localised effects in the brain corresponding to a change in the experimental condition) or \textit{reverse inference} (\emph{e.g.}, inferring the cognitive function from brain activation patterns, which is typically done using single subjects linear statistical models). More recently, it has been suggested that brain decoding (\emph{e.g.}, \textit{predicting} an experimental condition from patterns of brain activation) is a more formal way to perform reverse inference \cite{Poldrack2011}, as it allows to identify brain structures that are selectively engaged during a specific cognitive task. A few works have tested large-scale brain decoding using classic machine learning or deep learning paradigms applied to fMRI data from different subjects and tasks \cite{Varoquaux2018,Schulz2020}. Some of these studies have taken a step forward and tested graph convolutional networks to decode fMRI brain activity by also taking in account brain structural \cite{Bontonou2020} or functional \cite{zhang2021functional} connectivity, exploiting the promising framework of Graph Signal Processing (GSP).
  
  GSP is a mathematical framework that aims at extending classical Fourier analysis to irregular domains represented by graphs. In particular, notions of graph frequencies, modes, and associated operators such as convolutions or filtering can then be defined. In the context of fMRI decoding, the use of GSP-based methods also brings forward interpretability questions.
  
  In this work, we are interested in better understanding what are the graph frequencies that are the most useful to decode fMRI signals. To this end, we introduce a neural network architecture called \textit{spectral ResNet} in which graph frequencies are used to define convolutions. Using a pruning technique, we dynamically identify which frequencies are the most useful to decode fMRI signals. We show that these frequencies are robust to changes in the datasets and the considered architectures. We perform experiments with both structural or functional graphs.
    
    
\section{Related Work}
\label{sec:related}
   
   In this document, tensors and matrices are noted in bold uppercase, and vectors in bold lowercase. We access their entries using Python-style indexing, where $:$ denotes all entries along the corresponding dimension. Constants are noted in uppercase, variables in lowercase, and sets in calligraphic. 
    
    GSP generalizes Fourier's approach to signals evolving on irregular structures by providing an adapted spectral space to decompose them in meaningful frequencies \cite{shuman2013emerging, ortega2018graph}. In this framework, we consider a weighted and undirected graph $\graphG = \tuple{\setV, \setE, \funW}$ with vertices $\setV = \set{\varV_1, \dots, \varV_\cstN}$ of cardinal $\card\setV = \cstN$, edges $\setE \subset \setV \times \setV$, and a weighting function $\funW : \setE \mapsto \R$. Such a graph can be equivalently represented by its weights matrix $\matW \in \R^{\cstN \times \cstN}$ such that $\entry\matW{\varI, \varJ} = \funW\left(\set{\varV_\varI, \varV_\varJ}\right)$ if $\set{\varV_\varI, \varV_\varJ} \in \setE$ and $0$ otherwise. Additionally, we note $\matD \in \R^{\cstN \times \cstN}$ the degrees matrix of $\graphG$, such that $\entry\matD{\varI, \varJ} = \sum_{\varK = 1}^{\cstN} \entry\matW{\varI, \varK}$ if $\varI = \varJ$ and $0$ otherwise. From these two matrices, we can compute the normalized Laplacian matrix $\matL = \matI\cstN - \matD^{-1/2} \matW \matD^{-1/2}$ of $\graphG$, where $\matI\cstN$ is the identity matrix of dimension $\cstN$. Since $\matL$ is real and symmetric, it can be diagonalized as $\matL = \matU \matLambda \matU^\top$, where $\matU$ is a matrix of orthonormal vectors associated with eigenvalues forming the diagonal matrix $\matLambda$, sorted in increasing order. These eigenvalues are analogous to frequencies in Fourier analysis, and are called \textit{graph frequencies}.
    
    A signal $\vecX \in \R^\cstN$ on $\graphG$ is an observation on each of its vertices. Its Graph Fourier Transform $\vecXs = \gft(\vecX) = \matU^\top \vecX$ shows the various contributions of eigenvalues of $\matL$ in $\vecX$. Its inverse $\vecX = \igft(\vecXs) = \matU \vecXs$ transforms a graph spectrum into a graph signal. Following the usual setting of Fourier Analysis, convolution can be defined as entrywise multiplication $\odot$ in the spectral space. In more details, denote $\vecH \in \R^\cstN$ a filter, then convolution is obtained as:
    \begin{equation}
        \vecX * \vecH = \igft\left(\gft(\vecX) \odot \gft(\vecH)\right)\;.
    \end{equation}

 In previous works applying GSP to decode brain activity~\cite{Bontonou2020,zhang2021functional}, the focus has been mainly on increasing generalization accuracy. Here we aim at investigating which graph frequencies are the most relevant for brain decoding, and compare results for functional and structural graphs. We also assess the reliability of our results by applying the same analysis to two large scale datasets and testing different deep learning architectures. More specifically we address the following questions:
 \begin{enumerate}
     \item Which graph frequencies are the most relevant to predict the experimental task from corresponding fMRI activation maps?
     \item Are these frequencies consistent for different underlying graphs (functional vs. structural) and deep learning architectures?
     \item Does selecting specific graph frequency bands increase brain decoding performance?
 \end{enumerate}

    
\section{Methodology}
\label{sec:methodology}

In this work, we aim at exploiting the graph on which input signals are defined to improve the classification of these signals and its interpretability in terms of relevant brain spatial patterns. The core idea of the proposed approach is to select by a learning process the appropriate graph frequencies within the signals. Beyond the performance aspect, this also provides an understandable representation of the signal features that are used for training a model. To reach this goal, we build a spectral model around a $\gspConv$ layer, defined below.

\subsection{GSPConv layer}

A  $\gspConv$ layer is defined using a graph $\graphG$ with $\cstN$ vertices. It takes as input a matrix $\matX \in \R^{\cstN \times \varCin}$ seen as a graph signal with $\varCin$ channels and outputs a matrix $\matZ \in \R^{\cstN \times \varCout}$ seen as a graph signal with $\varCout$ channels. It contains a weight tensor $\tenW \in \R^{\cstN \times \varCin \times \varCout}$ and its mathematical function is:
\begin{equation}
    \gspConv\limits_{\varCin, \varCout}(\matX) = \sigma\left(\igft\left( \gft\left( \matX \right) \otimes \tenW\right)\right) \;,
\end{equation}
where $\gft$ and $\igft$ are applied to each channel independently, $\sigma$ is a nonlinear activation function, and $\otimes$ is such that $\entry{(\matXs \otimes \tenW)}{\varL, \varC} = \sum_{\varC'=1}^{\varCin}{\entry\matXs{\varL, \varC'} \entry\tenW{\varL, \varC', \varC}}$.

In other words, a $\gspConv$ is a natural extension of usual convolutional layers used in regular 1d or 2d domains.

\subsection{Considered architectures}

\begin{figure*}
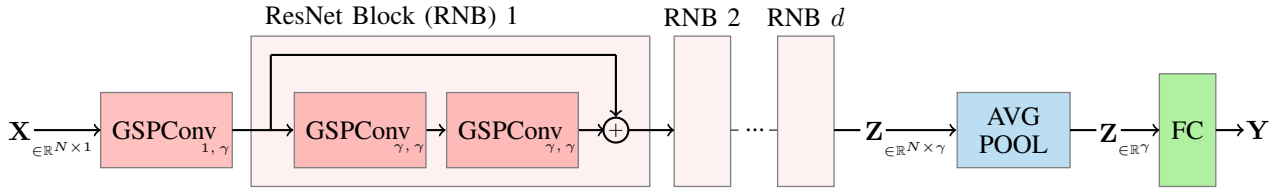

\ctikzfig{tikz/paper_figure}
\caption{Illustration of the Spectral ResNet model introduced in this paper. The input is seen as a graph signal with 1 channel, it is first embedded as a graph signal with $\gamma$ channels using a $\gspConv$ layer. Then it goes through $d$ ResNet blocks consisting of the sum between a shortcut path and a sequential path containing 2 $\gspConv$ layers. The output is seen as a graph signal with $\gamma$ channels. The values are averaged over the graph before being fed to a classical logistic regression.}
\label{fig:spectralRN}
\end{figure*}

Using the $\gspConv$ layer, we build a spectral ResNet, inspired by regular ResNets~\cite{he2015deep}. Such a model consists in assembling a first $\gspConv$ layer, then spectral ResNet blocks (c.f. Figure~\ref{fig:spectralRN}), and finally a fully connected layer applied on globally averaged values over the graph. This architecture is summarized in Figure~\ref{fig:spectralRN}. It is parametrized by a depth $\varD$, and a number of channels $\gamma$. It contains a total number of trainable parameters $\Order(\gamma \cstN + 2 \varD \gamma^2 \cstN + \gamma \cstC)$ where $\cstC$ is the number of classes in the considered problem.

For comparison purposes, we also propose an architecture based on a standard Multi-Layer Perceptron (MLP) which input domain is the graph frequency domain. For simplicity, we use a fixed number $\varH$ of hidden neurons in each layer and a depth $\varD$. Such a model contains a total of  $\Order(\varH \cstN + h^2 \varD + \varH \cstC)$ trainable parameters.

\subsection{Using pruning to identify important graph frequencies}

In the field of classification, pruning~\cite{han2015learning} is a very active topic of research that aims at identifying unimportant portions of deep learning architectures and remove them while maintaining a steady level of accuracy. The motivation is usually to reduce the number of parameters, hence the memory usage and the number of computations, resulting in lightweight models that can better fit constrained environments (\emph{e.g.}, edge computing). In this work, our aim is to adapt pruning in order to automatically identify the least important graph frequencies in our models, then prune them.

A pruning method usually combines multiple ingredients: 1- a pruning criterion which associates a subset of parameters in the considered architecture with an importance score, 2- a pruning technique which is typically deployed onto the parameters with the lowest importance score and 3- a pruning schedule which defines how the pruning ingredients interact with the global learning schedule of the considered architecture. We choose to implement Selective Weight Decay (SWD)~\cite{tessier2020rethinking} because it has the asset of allowing any subset of frequencies to be removed at any time during the training procedure, avoiding harsh sudden effects of other methods, while being state-of-the-art in competitive vision benchmarks.

In short, SWD consists of the following: at each step of the learning process, we aggregate all weights associated with a particular graph frequency $\varL$ into a linearized vector. For MLP, only the first layer of the architecture is concerned, whereas for Spectral ResNets, we combine all layers weight tensors $\entry\tenW{\varL, \cdot, \cdot}$. We then compute the magnitude of obtained vectors, and sort them in descending order. The $\cstK$ first vectors of weights are left untouched, and the remaining ones are applied an additional weight decay. This additional weight decay grows exponentially during the learning procedure, starting at a very low value $\alpha_{\min}$ so that the effect of SWD is almost unnoticeable, and finishing at a value $\alpha_{\max}$ so large that it boils down to nullifying the considered values. Once the procedure is finished, we identify the least important graph frequencies and remove them from the model, then we retrain the remaining parameters using the same scheduler as for baseline architectures, what is referred to as LR-rewinding~\cite{renda2019comparing} in the literature.

Note that pruning graph frequencies has a very different impact on the total number of parameters of considered models. For MLP, it reduces the parameters to $\Order(\varH \cstK + h^2 \varD + \varH \cstC)$ whereas for Spectral ResNets, it reduces to $\Order(\gamma \cstK + 2 \varD \gamma^2 \cstK + \gamma \cstC)$, resulting in a way more dramatic effect.

\subsection{Neuroimaging Data and Graphs}
Two datasets of spatial brain activation maps (obtained using general linear models of functional MRI signals~\cite{penny2011statistical}) were fetched from Neurovault~\cite{gorgolewski2015neurovault}: the second release of IBC~\cite{pinho2020individual} (Neurovault collection 6618), consisting of 13 subjects with many experimental tasks, and 788 subjects out of the 900 subjects release of the Human Connectome Project~\cite{van2012human} (Neurovault collection 4337). 
As labels for classification, we used the 24 tasks for IBC, and the 7 tasks for HCP. Here, we consider a large spectrum of experimental conditions ranging from motor to memory and social tasks. The datasets were split in training, validation and test splits. Following the usual supervised learning pipeline, training data was used to tune model parameters, validation data to monitor generalisation accuracy without parameter tuning, and the final accuracy reported is measured on the test set, using the model with the best validation accuracy. For HCP, 70~\% of subjects were used for training, 15~\% for validation and 15~\% for test. For IBC eight subjects were used for training, three for validation and two for test. The obtained splits were all balanced. 

In this work, we considered two brain graphs to build spectral filters. The first is an average structural graph estimated from diffusion weighted images of the HCP dataset (56 healthy subjects). Graph nodes correspond to the 360 regions of the Glasser atlas~\cite{Glasser2017} while graph edges are a measure of structural connectivity strength, as described in~\cite{Preti}. We also considered another consensus graph, estimated from resting state fMRI data of 1080 HCP healthy subjects. Also, for the functional graph, the nodes correspond to the 360 regions of the Glasser atlas, while edges are a measure of the Pearson correlation between fMRI time-series during resting state~\cite{zhang2021functional}. In order to ensure that the brain graphs had good spectral properties while remaining connected, binarized $k$-nearest-neighbor graphs were built from the original structural and functional graphs by connecting each node to its $k=8$ neighbors with strongest connectivity. All the considered models and datasets are available at our Github\footnote{https://github.com/elouayas/gspconv}.
    
\section{Experiments}
\label{sec:experiments}

We conducted a set of experiments to answer the series of questions established in Section \ref{sec:related}.

\subsection{Experimental details}

As a first step, we ran numerous experiments to find good hyperparameters for our considered architectures. The optimal hyperparameters are displayed in Table~\ref{hyperparams}. Note that ResNets contain way more parameters than their MLP counterpart, but as mentioned in Section~\ref{sec:methodology}, the number of these parameters dramatically reduces as we restrain the number of considered graph frequencies. We used batch normalization at the output of each $\gspConv$ layer and rectified linear units as activation functions. We standardized the datasets (so that data is centered and unit-normed) before training and used mixup~\cite{zhang2018mixup} during training. We report the averages and 95\% confidence intervals over multiple runs.

We observe that there is no model that is universally better than the others in terms of raw performance. Yet the structural version of ResNet performs very poorly on IBC compared to the alternatives, suggesting the structural graph is less adapted to the considered datasets than its functional counterpart.

\begin{table}
\centering
\caption{Hyperparameters and associated performance for our baseline architectures. ResNet S refers to the structural graph and ResNet F to the functional graph.}
\label{hyperparams}
\begin{tabular}{ | c | c | c | c | c|c|}
  \hline
  \textbf{model} & \textbf{depth} & \textbf{$\varH/\gamma$} & \textbf{parameters} & \textbf{IBC} & \textbf{HCP} \\ \hline
  MLP & 3 & 10 & 2.5M & 61.6$\pm0.5$ & 97.4$\pm0.03$ \\ \hline
  ResNet F & 4 & 7 & 47M & 66.3$\pm0.5$ & 96.9$\pm0.06$ \\ \hline
  ResNet S & 4 & 7 & 47M & 55.1$\pm0.5$ & 97.0$\pm0.08$ \\ \hline
\end{tabular}
\end{table}

%
%
%
%
%
%

\begin{figure*}
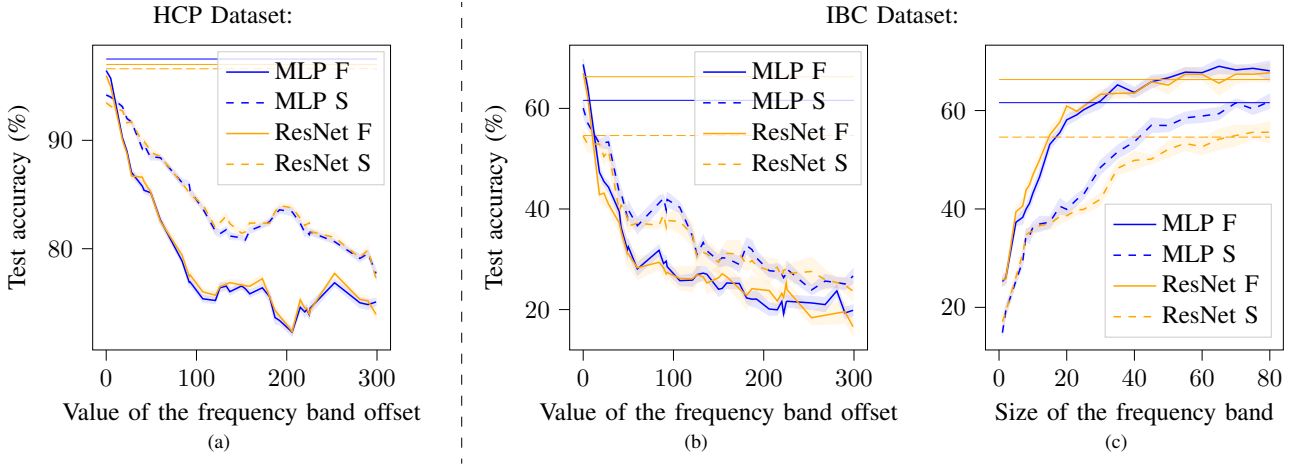

    \scalebox{0.95}{
        \begin{tabular}{c:c}
        HCP Dataset: & IBC Dataset:\\
        \subfigure[]{\input{figures/hcp_filtering_60.tex} \label{fig:submovingbandandchoosingbandwidtha}} &
        \subfigure[]{\input{figures/ibc_filtering_60.tex} \label{fig:submovingbandandchoosingbandwidthb}} 
        \subfigure[]{
\begin{tikzpicture}

\definecolor{darkgray176}{RGB}{176,176,176}
\definecolor{lightgray204}{RGB}{204,204,204}
\definecolor{orange}{RGB}{255,165,0}

\begin{axis}[
legend cell align={left},
legend style={
  fill opacity=0.8,
  draw opacity=1,
  text opacity=1,
  at={(0.40,0.48)},
  anchor=north west,
  draw=lightgray204
},
tick align=outside,
tick pos=left,
x grid style={darkgray176},
xlabel={Size of the frequency band},
xmin=-4, xmax=84,
xtick style={color=black},
y grid style={darkgray176},
ylabel={},
ymin=11.3417218315695, ymax=73.2870416166534,
ytick style={color=black},
width = 5.75 cm,
height = 5.85 cm
]
\path [draw=blue, fill=blue, opacity=0.1]
(axis cs:1,26.220770170634)
--(axis cs:1,24.2970292862802)
--(axis cs:2,24.8130205991078)
--(axis cs:5,36.5063325871653)
--(axis cs:7,36.9404283577491)
--(axis cs:8,39.1031960143529)
--(axis cs:9,39.8988167852557)
--(axis cs:10,41.9501628925)
--(axis cs:12,45.1749511694168)
--(axis cs:15,52.0135705185451)
--(axis cs:18,54.5944833847291)
--(axis cs:20,57.0162390752088)
--(axis cs:23,58.1421151149486)
--(axis cs:25,59.0699942329526)
--(axis cs:30,60.3989650651442)
--(axis cs:35,64.0641887353152)
--(axis cs:40,62.5943710215527)
--(axis cs:45,64.5477178434896)
--(axis cs:50,65.8363639612422)
--(axis cs:55,66.8616675156417)
--(axis cs:60,66.6469902511029)
--(axis cs:65,67.5221843748687)
--(axis cs:70,67.2709900932221)
--(axis cs:75,67.730720948053)
--(axis cs:80,66.0782662714562)
--(axis cs:80,70.0706029251177)
--(axis cs:80,70.0706029251177)
--(axis cs:75,69.4213858138793)
--(axis cs:70,69.23386665218)
--(axis cs:65,70.4713452627859)
--(axis cs:60,68.7575409098875)
--(axis cs:55,68.7370388887264)
--(axis cs:50,67.4646096130783)
--(axis cs:45,67.1351326763265)
--(axis cs:40,64.7195464563729)
--(axis cs:35,66.3888876272946)
--(axis cs:30,63.4845310603949)
--(axis cs:25,61.2536319674055)
--(axis cs:23,60.1426746062224)
--(axis cs:20,59.1973536130338)
--(axis cs:18,56.5058414049539)
--(axis cs:15,54.2323835181676)
--(axis cs:12,47.9318459694331)
--(axis cs:10,44.457604641756)
--(axis cs:9,42.3665550936861)
--(axis cs:8,40.8644430345413)
--(axis cs:7,39.6615148172489)
--(axis cs:5,38.0891374439372)
--(axis cs:2,27.1934519487412)
--(axis cs:1,26.220770170634)
--cycle;

\path [draw=blue, fill=blue, opacity=0.1]
(axis cs:1,15.5513197442252)
--(axis cs:1,14.1574181854369)
--(axis cs:2,18.2807167472435)
--(axis cs:5,24.7302553663659)
--(axis cs:7,28.3389892700863)
--(axis cs:8,33.4962848984737)
--(axis cs:9,32.9663946582316)
--(axis cs:10,35.2514178868311)
--(axis cs:12,35.3956229480716)
--(axis cs:15,36.1339364179584)
--(axis cs:18,38.9549857914066)
--(axis cs:20,38.7010144720337)
--(axis cs:23,40.7930979080973)
--(axis cs:25,42.019509323255)
--(axis cs:30,47.0451370074171)
--(axis cs:35,50.1914439914543)
--(axis cs:40,52.2044203572057)
--(axis cs:45,55.5193439449322)
--(axis cs:50,55.6942968625149)
--(axis cs:55,57.3500882093087)
--(axis cs:60,57.571628325967)
--(axis cs:65,58.1174781862603)
--(axis cs:70,60.7880214769791)
--(axis cs:75,59.1946173072384)
--(axis cs:80,60.4251724759817)
--(axis cs:80,63.3612372199615)
--(axis cs:80,63.3612372199615)
--(axis cs:75,61.4526335040206)
--(axis cs:70,62.3511366129766)
--(axis cs:65,60.620387102792)
--(axis cs:60,60.1953631303827)
--(axis cs:55,59.672566006186)
--(axis cs:50,58.2215614062229)
--(axis cs:45,58.4936035349516)
--(axis cs:40,55.1094951338984)
--(axis cs:35,52.9477140985014)
--(axis cs:30,49.8804303811175)
--(axis cs:25,43.9351838352126)
--(axis cs:23,43.0903974227132)
--(axis cs:20,41.1371744146087)
--(axis cs:18,42.1129760603491)
--(axis cs:15,38.493896153707)
--(axis cs:12,38.4878731774676)
--(axis cs:10,37.2728547139786)
--(axis cs:9,36.1598199254832)
--(axis cs:8,35.8888292150161)
--(axis cs:7,30.3988752242374)
--(axis cs:5,26.046444415052)
--(axis cs:2,20.3276981808749)
--(axis cs:1,15.5513197442252)
--cycle;

\path [draw=orange, fill=orange, opacity=0.1]
(axis cs:1,26.7274497787314)
--(axis cs:1,24.3890563805741)
--(axis cs:2,24.4118170249397)
--(axis cs:5,38.2622582706593)
--(axis cs:7,39.191357424652)
--(axis cs:8,42.7249271725723)
--(axis cs:9,43.6200370109106)
--(axis cs:10,45.809964955436)
--(axis cs:12,48.7695235662827)
--(axis cs:15,53.607889112574)
--(axis cs:18,56.6306965382152)
--(axis cs:20,59.7881808772591)
--(axis cs:23,58.9745785752175)
--(axis cs:25,59.7431894834705)
--(axis cs:30,62.2958464028465)
--(axis cs:35,62.4591682835929)
--(axis cs:40,62.7594651816748)
--(axis cs:45,64.8402334247786)
--(axis cs:50,63.3743238863341)
--(axis cs:55,65.9652156192572)
--(axis cs:60,66.2524282329512)
--(axis cs:65,63.0508445203343)
--(axis cs:70,65.5363541866709)
--(axis cs:75,65.9140047426726)
--(axis cs:80,65.7506979509251)
--(axis cs:80,69.5891105131252)
--(axis cs:80,69.5891105131252)
--(axis cs:75,68.8432779912447)
--(axis cs:70,69.2694729541372)
--(axis cs:65,68.0656601488552)
--(axis cs:60,68.4563100940752)
--(axis cs:55,68.792068306753)
--(axis cs:50,66.9654821935304)
--(axis cs:45,66.8587969745439)
--(axis cs:40,64.3279145123102)
--(axis cs:35,64.5311249807962)
--(axis cs:30,64.3060963747871)
--(axis cs:25,62.0043865148358)
--(axis cs:23,60.6856177291038)
--(axis cs:20,62.0564791664574)
--(axis cs:18,58.7091095416493)
--(axis cs:15,56.4891997012078)
--(axis cs:12,51.2304776258102)
--(axis cs:10,48.0735315674674)
--(axis cs:9,46.1372445309137)
--(axis cs:8,44.9935196782044)
--(axis cs:7,42.0222360583192)
--(axis cs:5,40.5726944413998)
--(axis cs:2,27.2386689674919)
--(axis cs:1,26.7274497787314)
--cycle;

\path [draw=orange, fill=orange, opacity=0.1]
(axis cs:1,18.9649428795403)
--(axis cs:1,15.2098150183135)
--(axis cs:2,17.9322116008093)
--(axis cs:5,24.9020244809137)
--(axis cs:7,29.6894171900058)
--(axis cs:8,33.5660881424134)
--(axis cs:9,34.1667714371921)
--(axis cs:10,34.3055457435973)
--(axis cs:12,35.9132328692987)
--(axis cs:15,35.7619939223005)
--(axis cs:18,36.9288329492661)
--(axis cs:20,37.5797191859765)
--(axis cs:23,38.801229992872)
--(axis cs:25,38.9064910847133)
--(axis cs:30,40.7065226914128)
--(axis cs:35,46.963970821393)
--(axis cs:40,48.079334038044)
--(axis cs:45,49.2350694214542)
--(axis cs:50,50.4521917114511)
--(axis cs:55,51.8617111281861)
--(axis cs:60,51.047037311288)
--(axis cs:65,52.119031764398)
--(axis cs:70,52.7974298529207)
--(axis cs:75,54.2080280970028)
--(axis cs:80,53.3984900401179)
--(axis cs:80,57.7665590359624)
--(axis cs:80,57.7665590359624)
--(axis cs:75,56.9570221711705)
--(axis cs:70,57.0569393105925)
--(axis cs:65,56.2304840313097)
--(axis cs:60,54.2442243616856)
--(axis cs:55,54.7402316494475)
--(axis cs:50,53.7711091031775)
--(axis cs:45,51.007648440675)
--(axis cs:40,51.6779480998267)
--(axis cs:35,49.5408843345518)
--(axis cs:30,43.2740608571331)
--(axis cs:25,40.8993348640019)
--(axis cs:23,40.7618774976676)
--(axis cs:20,39.6532913444953)
--(axis cs:18,40.0614585269836)
--(axis cs:15,38.2185890301989)
--(axis cs:12,37.2906514938757)
--(axis cs:10,36.6653286851517)
--(axis cs:9,36.3186666712521)
--(axis cs:8,35.9484755372817)
--(axis cs:7,32.1066996389127)
--(axis cs:5,27.3795286683096)
--(axis cs:2,20.3202159771632)
--(axis cs:1,18.9649428795403)
--cycle;

\addplot [semithick, blue]
table {%
1 25.2588997284571
2 26.0032362739245
5 37.2977350155512
7 38.300971587499
8 39.9838195244471
9 41.1326859394709
10 43.203883767128
12 46.5533985694249
15 53.1229770183563
18 55.5501623948415
20 58.1067963441213
23 59.1423948605855
25 60.161813100179
30 61.9417480627696
35 65.2265381813049
40 63.6569587389628
45 65.841425259908
50 66.6504867871602
55 67.799353202184
60 67.7022655804952
65 68.9967648188273
70 68.252428372701
75 68.5760533809662
80 68.0744345982869
};
\addlegendentry{MLP F}
\addplot [semithick, blue, dashed]
table {%
1 14.854368964831
2 19.3042074640592
5 25.3883498907089
7 29.3689322471619
8 34.6925570567449
9 34.5631072918574
10 36.2621363004049
12 36.9417480627696
15 37.3139162858327
18 40.5339809258779
20 39.9190944433212
23 41.9417476654053
25 42.9773465792338
30 48.4627836942673
35 51.5695790449778
40 53.6569577455521
45 57.0064737399419
50 56.9579291343689
55 58.5113271077474
60 58.8834957281749
65 59.3689326445262
70 61.5695790449778
75 60.3236254056295
80 61.8932048479716
};
\addlegendentry{MLP S}
\addplot [semithick, orange]
table {%
1 25.5582530796528
2 25.8252429962158
5 39.4174763560295
7 40.6067967414856
8 43.8592234253883
9 44.8786407709122
10 46.9417482614517
12 50.0000005960464
15 55.0485444068909
18 57.6699030399322
20 60.9223300218582
23 59.8300981521607
25 60.8737879991531
30 63.3009713888168
35 63.4951466321945
40 63.5436898469925
45 65.8495151996613
50 65.1699030399323
55 67.3786419630051
60 67.3543691635132
65 65.5582523345947
70 67.4029135704041
75 67.3786413669586
80 67.6699042320251
};
\addlegendentry{ResNet F}
\addplot [semithick, orange, dashed]
table {%
1 17.0873789489269
2 19.1262137889862
5 26.1407765746117
7 30.8980584144592
8 34.7572818398476
9 35.2427190542221
10 35.4854372143745
12 36.6019421815872
15 36.9902914762497
18 38.4951457381248
20 38.6165052652359
23 39.7815537452698
25 39.9029129743576
30 41.9902917742729
35 48.2524275779724
40 49.8786410689354
45 50.1213589310646
50 52.1116504073143
55 53.3009713888168
60 52.6456308364868
65 54.1747578978539
70 54.9271845817566
75 55.5825251340866
80 55.5825245380402
};
\addlegendentry{ResNet S}
\path [draw=blue]
(axis cs:0,61.6)
--(axis cs:80,61.6);

\path [draw=orange]
(axis cs:0,66.3)
--(axis cs:80,66.3);

\path [draw=orange, dash pattern=on 3.7pt off 1.6pt]
(axis cs:0,54.6)
--(axis cs:80,54.6);

\end{axis}

\end{tikzpicture} \label{fig:submovingbandandchoosingbandwidthc}}
        \end{tabular}
    }
    \caption{(a)(b) Evolution of the accuracy of various architectures/datasets as a function of the graph frequency band offset. (c) Evolution of the accuracy of various architectures on the IBC dataset as a function of the graph frequency bandwidth, when the offset is 0. The horizontal lines correspond to the baseline accuracy. Shaded areas correspond to 95\% confidence intervals.}
    \label{fig:movingbandandchoosingbandwidth}
\end{figure*}

\subsection{Impact of graph frequencies}

To set up a baseline for the pruning methodology, we first evaluate the performance of the proposed architectures when we apply a band selection of the graph frequencies. To this end, we first identify which parts of the spectrum seem the most useful for decoding fMRI signals. The results are depicted in Figures~\ref{fig:submovingbandandchoosingbandwidtha} and \ref{fig:submovingbandandchoosingbandwidthb}. We fixed a bandwidth of 60 frequencies and varied the offset, which corresponds to the x axis. We observe that no matter the architecture nor the graph, best results are obtained by keeping the low graph frequencies.

In Figure~\ref{fig:submovingbandandchoosingbandwidthc}, we vary the bandwidth when the offset is set to 0. We observe that keeping more than 60 frequencies does not benefit accuracy by a large margin. We again observe that the functional graph leads to better results overall than the structural one. Interestingly, in some cases keeping a portion of the frequencies leads to better results than the baselines.

%

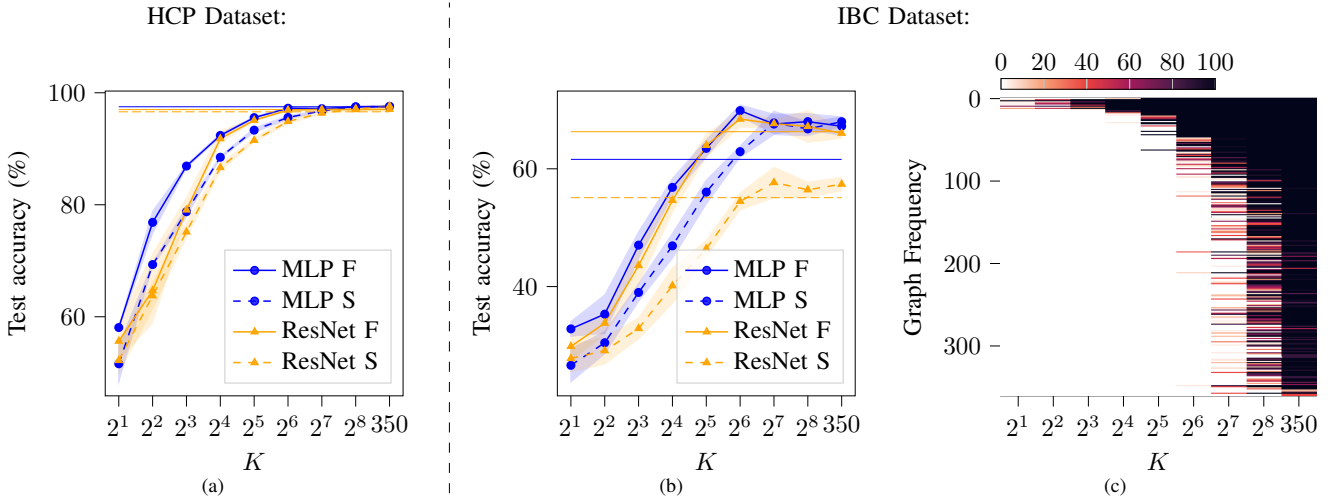
\begin{figure*}
    \scalebox{0.95}{
        \begin{tabular}{c:c}
        HCP Dataset: & IBC Dataset:\\
        \subfigure[]{
\begin{tikzpicture}

\definecolor{darkgray176}{RGB}{176,176,176}
\definecolor{lightgray204}{RGB}{204,204,204}
\definecolor{orange}{RGB}{255,165,0}

\begin{axis}[
legend cell align={left},
legend style={
  fill opacity=0.8,
  draw opacity=1,
  text opacity=1,
  at={(0.40,0.48)},
  anchor=north west,
  draw=lightgray204
},
tick align=outside,
tick pos=left,
x grid style={darkgray176},
xlabel={$K$},
xmin=-0.4, xmax=8.4,
xtick style={color=black},
xtick={0,1,2,3,4,5,6,7,8},
xticklabels={$2^1$,$2^2$,$2^3$,$2^4$,$2^5$,$2^6$,$2^7$,$2^8$,$350$},
y grid style={darkgray176},
ylabel={Test accuracy (\%)},
ymin=45.8767139629676, ymax=100.305710480547,
ytick style={color=black},
width = 5.75 cm,
height = 5.85 cm
]
\path [draw=blue, fill=blue, opacity=0.15]
(axis cs:0,58.3407917471767)
--(axis cs:0,57.8093037156224)
--(axis cs:1,74.8603521566039)
--(axis cs:2,86.5838465856453)
--(axis cs:3,92.0838321349403)
--(axis cs:4,95.362489537138)
--(axis cs:5,97.0700031838349)
--(axis cs:6,96.9491843699197)
--(axis cs:7,97.3713433934228)
--(axis cs:8,97.3690591526946)
--(axis cs:8,97.7153672503511)
--(axis cs:8,97.7153672503511)
--(axis cs:7,97.6005136775)
--(axis cs:6,97.3847742558737)
--(axis cs:5,97.4099188834258)
--(axis cs:4,95.759461804491)
--(axis cs:3,92.6816462376336)
--(axis cs:2,87.2397937608818)
--(axis cs:1,78.8582218904847)
--(axis cs:0,58.3407917471767)
--cycle;

\path [draw=blue, fill=blue, opacity=0.15]
(axis cs:0,54.8537426080731)
--(axis cs:0,48.3507592592212)
--(axis cs:1,68.9793617425524)
--(axis cs:2,77.9110951337968)
--(axis cs:3,88.288213252264)
--(axis cs:4,92.9324241606397)
--(axis cs:5,95.2811481157063)
--(axis cs:6,96.6531089935926)
--(axis cs:7,97.2768373176635)
--(axis cs:8,97.4736299810897)
--(axis cs:8,97.6595798434723)
--(axis cs:8,97.6595798434723)
--(axis cs:7,97.6800136878906)
--(axis cs:6,96.9003601874682)
--(axis cs:5,95.9083460172893)
--(axis cs:4,93.7167314561205)
--(axis cs:3,88.687397003931)
--(axis cs:2,79.6123545255507)
--(axis cs:1,69.662287742082)
--(axis cs:0,54.8537426080731)
--cycle;

\path [draw=orange, fill=orange, opacity=0.15]
(axis cs:0,58.9096034825863)
--(axis cs:0,52.3774505793034)
--(axis cs:1,58.855168884876)
--(axis cs:2,78.1505898970063)
--(axis cs:3,91.0064676528204)
--(axis cs:4,94.8902599915649)
--(axis cs:5,96.6838800477073)
--(axis cs:6,96.9289549498)
--(axis cs:7,96.7883574427464)
--(axis cs:8,96.7163465119029)
--(axis cs:8,97.3699595832203)
--(axis cs:8,97.3699595832203)
--(axis cs:7,97.4174309651668)
--(axis cs:6,97.1443659253678)
--(axis cs:5,97.1365327406838)
--(axis cs:4,95.3685142206707)
--(axis cs:3,93.2324605673946)
--(axis cs:2,81.2679752603063)
--(axis cs:1,71.3259689506594)
--(axis cs:0,58.9096034825863)
--cycle;

\path [draw=orange, fill=orange, opacity=0.15]
(axis cs:0,55.1733403191577)
--(axis cs:0,49.9104371749161)
--(axis cs:1,61.3593804706636)
--(axis cs:2,74.1593859981453)
--(axis cs:3,85.959926970558)
--(axis cs:4,91.2203679727925)
--(axis cs:5,94.4788856963867)
--(axis cs:6,96.2007230763579)
--(axis cs:7,97.0066056089174)
--(axis cs:8,97.4560662733964)
--(axis cs:8,97.8316651842937)
--(axis cs:8,97.8316651842937)
--(axis cs:7,97.454782503303)
--(axis cs:6,96.630909816581)
--(axis cs:5,95.6188766543055)
--(axis cs:4,91.8221163261841)
--(axis cs:3,87.3468335554696)
--(axis cs:2,76.8010095884347)
--(axis cs:1,66.1453092227873)
--(axis cs:0,55.1733403191577)
--cycle;

\addplot [semithick, blue, mark=*, mark size=1.5, mark options={solid}]
table {%
0 58.0750477313995
1 76.8592870235443
2 86.9118201732635
3 92.3827391862869
4 95.5609756708145
5 97.2399610336304
6 97.1669793128967
7 97.4859285354614
8 97.5422132015228
};
\addlegendentry{MLP F}
\addplot [semithick, blue, dashed, mark=*, mark size=1.5, mark options={solid}]
table {%
0 51.6022509336472
1 69.3208247423172
2 78.7617248296738
3 88.4878051280975
4 93.3245778083801
5 95.5947470664978
6 96.7767345905304
7 97.4784255027771
8 97.566604912281
};
\addlegendentry{MLP S}
\addplot [semithick, orange, mark=*, mark size=1.5, mark options={solid},mark=triangle*]
table {%
0 55.6435270309448
1 64.6829271316528
2 79.0694177150726
3 91.804878115654
4 95.0806760787964
5 96.9102063941956
6 97.0366604375839
7 97.0131325721741
8 97.0431530475616
};
\addlegendentry{ResNet F}
\addplot [semithick, orange, dashed, mark=*, mark size=1.5, mark options={solid},mark=triangle*]
table {%
0 52.1951234340668
1 63.7523448467255
2 75.151971578598
3 86.6491550207138
4 91.5009379386902
5 94.8893070220947
6 96.3752347230911
7 97.1632266044617
8 97.6285183429718
};
\addlegendentry{ResNet S}
\path [draw=blue]
(axis cs:0,97.5)
--(axis cs:8,97.5);

\path [draw=orange]
(axis cs:0,97)
--(axis cs:8,97);

\path [draw=orange, dash pattern=on 3.7pt off 1.6pt]
(axis cs:0,96.6)
--(axis cs:8,96.6);

\end{axis}

\end{tikzpicture} \label{fig:subpruningandkeptfrequenciesa}} &
        \subfigure[]{
\begin{tikzpicture}

\definecolor{darkgray176}{RGB}{176,176,176}
\definecolor{lightgray204}{RGB}{204,204,204}
\definecolor{orange}{RGB}{255,165,0}

\begin{axis}[
legend cell align={left},
legend style={
  fill opacity=0.8,
  draw opacity=1,
  text opacity=1,
  at={(0.40,0.48)},
  anchor=north west,
  draw=lightgray204
},
tick align=outside,
tick pos=left,
x grid style={darkgray176},
xlabel={$K$},
xmin=-0.4, xmax=8.4,
xtick style={color=black},
xtick={0,1,2,3,4,5,6,7,8},
xticklabels={$2^1$,$2^2$,$2^3$,$2^4$,$2^5$,$2^6$,$2^7$,$2^8$,$350$},
y grid style={darkgray176},
ylabel={Test accuracy (\%)},
ymin=21.445268932302, ymax=73.2027265248255,
ytick style={color=black},
width = 5.75 cm,
height = 5.85 cm
]
\path [draw=blue, fill=blue, opacity=0.15]
(axis cs:0,34.0724064178894)
--(axis cs:0,31.5586619309952)
--(axis cs:1,32.0484431908554)
--(axis cs:2,44.7590490488029)
--(axis cs:3,55.3380843625521)
--(axis cs:4,61.9970026426415)
--(axis cs:5,68.909961400814)
--(axis cs:6,65.9311814269787)
--(axis cs:7,66.6578177083687)
--(axis cs:8,65.6543431855935)
--(axis cs:8,68.8116778753502)
--(axis cs:8,68.8116778753502)
--(axis cs:7,69.3615993391319)
--(axis cs:6,69.262993816399)
--(axis cs:5,70.8501148160745)
--(axis cs:4,64.9447450227637)
--(axis cs:3,58.3512357248807)
--(axis cs:2,49.3186221214318)
--(axis cs:1,38.5826251580291)
--(axis cs:0,34.0724064178894)
--cycle;

\path [draw=blue, fill=blue, opacity=0.15]
(axis cs:0,29.4545466388961)
--(axis cs:0,23.7978806410531)
--(axis cs:1,28.5728379295965)
--(axis cs:2,37.5044063718055)
--(axis cs:3,44.7877208978949)
--(axis cs:4,53.9967001606414)
--(axis cs:5,61.9483189958156)
--(axis cs:6,65.8078789123117)
--(axis cs:7,65.5957931124719)
--(axis cs:8,67.2277686769254)
--(axis cs:8,68.7673779552691)
--(axis cs:8,68.7673779552691)
--(axis cs:7,67.947897330567)
--(axis cs:6,69.7792683159294)
--(axis cs:5,63.876926384586)
--(axis cs:4,58.0906789372971)
--(axis cs:3,49.0472306220713)
--(axis cs:2,40.5053039400842)
--(axis cs:1,32.3494920922617)
--(axis cs:0,29.4545466388961)
--cycle;

\path [draw=orange, fill=orange, opacity=0.15]
(axis cs:0,32.2503233351756)
--(axis cs:0,27.4292891225766)
--(axis cs:1,31.8981309219222)
--(axis cs:2,40.36919144441)
--(axis cs:3,52.6070140338033)
--(axis cs:4,62.071175227432)
--(axis cs:5,67.0913433474086)
--(axis cs:6,66.9747461040228)
--(axis cs:7,64.5024012353897)
--(axis cs:8,65.1698926217785)
--(axis cs:8,66.9271961920033)
--(axis cs:8,66.9271961920033)
--(axis cs:7,69.8665333959579)
--(axis cs:6,68.3708871166498)
--(axis cs:5,69.7746800977208)
--(axis cs:4,65.8899906775668)
--(axis cs:3,56.6648317837627)
--(axis cs:2,46.7958582277167)
--(axis cs:1,35.6746856884141)
--(axis cs:0,32.2503233351756)
--cycle;

\path [draw=orange, fill=orange, opacity=0.15]
(axis cs:0,30.0754373185646)
--(axis cs:0,25.5167960412491)
--(axis cs:1,26.8309629451369)
--(axis cs:2,31.1172058298852)
--(axis cs:3,37.1136406480239)
--(axis cs:4,45.0672023804095)
--(axis cs:5,53.0342810171645)
--(axis cs:6,55.0702140329522)
--(axis cs:7,55.1357614226643)
--(axis cs:8,56.2103108235253)
--(axis cs:8,58.5663877657997)
--(axis cs:8,58.5663877657997)
--(axis cs:7,57.7380235962566)
--(axis cs:6,60.2695920469123)
--(axis cs:5,55.9210607272584)
--(axis cs:4,47.7968764989469)
--(axis cs:3,43.2015558656441)
--(axis cs:2,34.6594927594397)
--(axis cs:1,31.3826311338807)
--(axis cs:0,30.0754373185646)
--cycle;

\addplot [semithick, blue, mark=*, mark size=1.5, mark options={solid}]
table {%
0 32.8155341744423
1 35.3155341744423
2 47.0388355851173
3 56.8446600437164
4 63.4708738327026
5 69.8800381084442
6 67.5970876216888
7 68.0097085237503
8 67.2330105304718
};
\addlegendentry{MLP F}
\addplot [semithick, blue, dashed, mark=*, mark size=1.5, mark options={solid}]
table {%
0 26.6262136399746
1 30.4611650109291
2 39.0048551559448
3 46.9174757599831
4 56.0436895489693
5 62.9126226902008
6 67.7935736141205
7 66.7718452215195
8 67.9975733160973
};
\addlegendentry{MLP S}
\addplot [semithick, orange, mark=*, mark size=1.5, mark options={solid},mark=triangle*]
table {%
0 29.8398062288761
1 33.7864083051681
2 43.5825248360634
3 54.635922908783
4 63.9805829524994
5 68.4330117225647
6 67.6728166103363
7 67.1844673156738
8 66.0485444068909
};
\addlegendentry{ResNet F}
\addplot [semithick, orange, dashed, mark=*, mark size=1.5, mark options={solid},mark=triangle*]
table {%
0 27.7961166799068
1 29.1067970395088
2 32.8883492946625
3 40.1456320285797
4 46.4320394396782
5 54.4776708722115
6 57.6699030399322
7 56.4368925094604
8 57.3883492946625
};
\addlegendentry{ResNet S}
\path [draw=blue]
(axis cs:0,61.6)
--(axis cs:8,61.6);

\path [draw=orange]
(axis cs:0,66.3)
--(axis cs:8,66.3);

\path [draw=orange, dash pattern=on 3.7pt off 1.6pt]
(axis cs:0,55.1)
--(axis cs:8,55.1);

\end{axis}

\end{tikzpicture} \label{fig:subpruningandkeptfrequenciesb}} 
        \subfigure[]{
\begin{tikzpicture}

\definecolor{darkgray176}{RGB}{176,176,176}

\begin{axis}[
colorbar horizontal,
colorbar style={ylabel={},at={(0.68,1.03)},anchor=south east,width=0.50*6cm,height=1.5mm,xticklabel pos=upper},
colormap={mymap}{[1pt]
  rgb(0pt)=(1,1,1);
  rgb(1pt)=(0.964705882352941,0.611764705882353,0.450980392156863);
  rgb(2pt)=(0.909803921568627,0.247058823529412,0.247058823529412);
  rgb(3pt)=(0.631372549019608,0.101960784313725,0.356862745098039);
  rgb(4pt)=(0.298039215686275,0.113725490196078,0.294117647058824);
  rgb(5pt)=(0.0117647058823529,0.0196078431372549,0.101960784313725)
},
point meta max=100,
point meta min=0,
tick align=outside,
tick pos=left,
x grid style={darkgray176},
xlabel={$K$},
xmin=0, xmax=9,
xtick style={color=black},
xtick={0.5,1.5,2.5,3.5,4.5,5.5,6.5,7.5,8.5},
xticklabels={$2^1$,$2^2$,$2^3$,$2^4$,$2^5$,$2^6$,$2^7$,$2^8$,$350$},
y dir=reverse,
y grid style={darkgray176},
ylabel={Graph Frequency},
ymin=-0.5, ymax=360.5,
ytick style={color=black},
width = 6 cm,
height = 5.75 cm
]
\addplot graphics [includegraphics cmd=\pgfimage,xmin=0, xmax=9, ymin=360.5, ymax=-0.5] {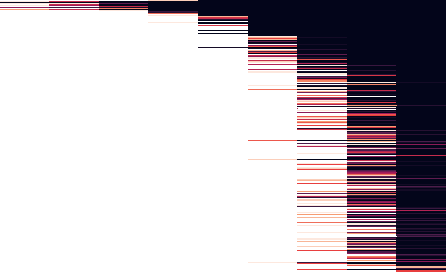};
\end{axis}

\end{tikzpicture} \label{fig:subpruningandkeptfrequenciesc}}
        \end{tabular}
    }
    \caption{
    (a)(b) Evolution of the accuracy of various architectures/datasets as a function of the number $K$ of kept frequencies obtained through our pruning methodology. The horizontal lines correspond to the baseline performance. Shaded areas correspond to 95\% confidence intervals. (c) Occurrences of each graph frequency for a fixed $K$ with the MLP Functional on IBC.}
    \label{fig:pruningandkeptfrequencies}
\end{figure*}


\subsection{Automated selection of optimal graph frequencies} 

We now test the proposed pruning methodology. Figures~\ref{fig:subpruningandkeptfrequenciesa} and \ref{fig:subpruningandkeptfrequenciesb} depict accuracy as a function of the number of kept frequencies $K$. Similarly to Figure~\ref{fig:submovingbandandchoosingbandwidthc}, we observe that the baseline cannot be outperformed with a limited number of selected frequencies. Again, the structural graph leads to poorer performance, suggesting that it is not only its lower spectrum that is not aligned with the task but that overall it offers poorer discrimination capabilities on our benchmarks.

Table~\ref{tab:filtervspruning} reports the best achieved gains with respect to our baselines, when using either the band or pruning method. We observe that pruning can consistently lead to improvements over all our experiments, whereas band selection degrades the performance in most considered scenarios. This table highlights the benefits of using the pruning methodology.


\begin{figure}
\includegraphics[scale=0.2]{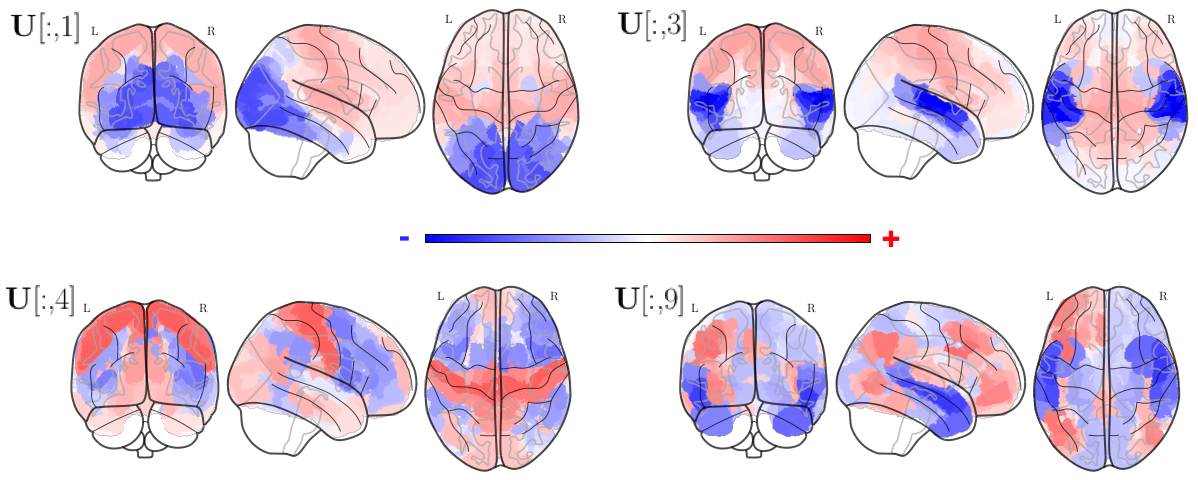}
\caption{Visualization of most frequently selected graph eigenvectors of the functional graph on the glass brain.}
\label{fig:brains}
\end{figure}

\begin{table}[]
\centering
\caption{Best gains obtained when using predetermined frequency bands or the pruning strategy on the considered architectures and datasets.}
\label{tab:filtervspruning}
\begin{tabular}{|c|cc|cc|}
\hline
\textbf{model}         & \multicolumn{2}{c|}{\textbf{IBC}}                                     & \multicolumn{2}{c|}{\textbf{HCP}}                                    \\ \hline
Size / K = 60 & \multicolumn{1}{c|}{Band}          & Pruning                 & \multicolumn{1}{c|}{Band}          & Pruning                \\ \hline
MLP F         & \multicolumn{1}{c|}{68.7$\pm$1.1} & \textbf{69.9$\pm$1.0}  & \multicolumn{1}{c|}{96.4$\pm$0.1} & \textbf{97.2$\pm$0.2} \\ \hline
MLP S         & \multicolumn{1}{c|}{60.0$\pm$1.5} & \textbf{62.9$\pm$1.0}  & \multicolumn{1}{c|}{94.2$\pm$0.1} & \textbf{95.6$\pm$0.3} \\ \hline
ResNet F      & \multicolumn{1}{c|}{67.0$\pm$1.7} & \textbf{68.4$\pm$1.3}  & \multicolumn{1}{c|}{96.0$\pm$0.2} & \textbf{96.9$\pm$0.2} \\ \hline
ResNet S      & \multicolumn{1}{c|}{54.5$\pm$1.4} & \textbf{54.7$\pm$1.4} & \multicolumn{1}{c|}{93.6 $\pm$0.2} & \textbf{94.9$\pm$0.4} \\ \hline
\end{tabular}
\end{table}

In Figure~\ref{fig:subpruningandkeptfrequenciesc}, we depict the frequencies selected using the pruning methodology, for various values of $K$, in the case of the spectral ResNets on the functional graph and IBC dataset. We perform 20 runs, and report the number of times each frequency was selected. Interestingly, we observe that the lower spectrum is overall preferred, even though it is not exactly a band of frequencies that is selected, explaining for the potential benefits over the predetermined frequency bands. We computed Intersection over Union (IoU) measures to quantify how reproducible these results are across datasets and architectures. Our IoU scores are typically above 50\%, suggesting high reproducibility.

In Figure~\ref{fig:brains}, we visualize on the glass brain the most frequently selected eigenvectors of the functional graph. These 4 eigenvectors -- $\entry{\matU}{:, 1}$, $\entry{\matU}{:, 3}$, $\entry{\matU}{:, 4}$, $\entry{\matU}{:, 9}$ associated with eigenvalues $\entry{\matLambda}{1,1}$, $\entry{\matLambda}{3,3}$, $\entry{\matLambda}{4,4}$, $\entry{\matLambda}{9,9}$ respectively -- are among those that appeared the most frequently when using pruning with $K=8$, under all configurations of models and datasets. Spatial distributions of largest positive and negative values of $\entry{\matU}{:, 1}$ and $\entry{\matU}{:, 3}$ correspond respectively to occipital and superior temporal brain areas, suggesting contributions of primary visual and auditory systems. $\entry{\matU}{:, 4}$ most positive values are confined to the motor cortex, while $\entry{\matU}{:, 9}$ includes medial contributions of the default mode network \cite{Smallwood2021}. These results show how the proposed methodology is able to identify meaningful graph frequencies (interpreted here as spatial patterns) for brain activity decoding. As we consider the graph frequencies common to all the tasks combined, general purpose brain patterns were found, consistent with the literature \cite{Mensch2021}. In future work we will investigate the potential of this methodology to identify spatial patterns characteristics of a specific decoded task or of a patient profile.

\section{Conclusions}
\label{sec:conclusions}

In this paper, we have introduced a simple deep learning architecture based on ResNets to process graph-based signals and applied it to fMRI decoding tasks.
Using a pruning methodology, we selected the most important graph frequencies and observed that keeping 60 out of the 360 initial frequencies could lead to improved performance.
Interestingly, the selected frequencies seem to be reproducible with other architectures and datasets.
We believe that this study could help designing more efficient and interpretable graph neural networks for this important domain of application.

\section{Acknowledgment}
\label{sec:acknowledgment}

We would like to thank Yu Zhang for providing the consensus functional graph. We also thank the region of Brittany for its support.
\\

    \bibliographystyle{IEEEtran}
    \bibliography{4_references}
\end{document}